\documentclass[aic]{iosart2x}

\usepackage[T1]{fontenc} 
\usepackage[utf8]{inputenc}

\usepackage{booktabs} 

\usepackage{graphicx}
\usepackage{xcolor}
\usepackage{amssymb}
\usepackage{amsmath}
\usepackage{fancyvrb}
\usepackage{subcaption}
\usepackage{hyperref}
\usepackage{marvosym}

\usepackage{footnote}
\makesavenoteenv{tabular}
\makesavenoteenv{table}

\newtheorem{example}{Example}
\newtheorem{definition}{Definition}

\pubyear{2019}
\volume{32}
\firstpage{271}
\lastpage{286}

\begin{document}
\makeatletter
\let\put@numberlines@box\relax
\makeatother

\begin{frontmatter}

\begin{center}
The Version of Record of this manuscript \\
has been published and is available in \\
\emph{AI Communications, vol. 32, no. 4, pp. 271-286 (2019)} \\
\url{https://doi.org/10.3233/AIC-190614}
\end{center}

\title{Stream-based perception for cognitive agents in mobile ecosystems}
\runningtitle{Stream-based perception for cognitive agents in mobile ecosystems}

\author[A]{\inits{JD}\fnms{Jeremias} \snm{Dötterl}\ead[label=e1]{jeremias.doetterl@hs-hannover.de}%
\thanks{Corresponding author. Hannover University of Applied Science and Arts, Computer Science Department, Ricklinger Stadtweg 120, 30459 Hannover, Germany. \printead{e1}.}},
\author[A]{\inits{RB}\fnms{Ralf} \snm{Bruns}\ead[label=e2]{ralf.bruns@hs-hannover.de}},
\author[A]{\inits{JD}\fnms{Jürgen} \snm{Dunkel}\ead[label=e3]{juergen.dunkel@hs-hannover.de}}
and
\author[B]{\inits{SO}\fnms{Sascha} \snm{Ossowski}\ead[label=e4]{sascha.ossowski@urjc.es}}
\runningauthor{J. Dötterl et al.}
\address[A]{Computer Science Department, \institution{Hannover University of Applied Sciences and Arts},
\cny{Germany}\printead[presep={\\}]{e1,e2,e3}}
\address[B]{Centre for Intelligent Information Technologies (CETINIA), \institution{Universidad Rey Juan Carlos}, Madrid, \cny{Spain}\printead[presep={\\}]{e4}}

\begin{abstract}
Cognitive agent abstractions can help to engineer intelligent systems across mobile devices.
On smartphones, the data obtained from onboard sensors can give valuable insights into the user's current situation.
Unfortunately, today's cognitive agent frameworks cannot cope well with the challenging characteristics of sensor data.
Sensor data is located on a low abstraction level and the individual data elements are not meaningful when observed in isolation.
In contrast, cognitive agents operate on high-level percepts and lack the means to effectively detect complex spatio-temporal patterns in sequences of multiple percepts.
In this paper, we present a stream-based perception approach that enables the agents to perceive meaningful situations in low-level sensor data streams.
We present a crowdshipping case study where autonomous, self-interested agents collaborate to deliver parcels to their destinations.
We show how situations derived from smartphone sensor data can trigger and guide auctions, which the agents use to reach agreements.
Experiments with real smartphone data demonstrate the benefits of stream-based agent perception.
\end{abstract}


\end{frontmatter}

\section{Introduction}
\label{sec:introduction}
In recent years, mobile devices have become mature computing platforms that are taking over an ever-increasing number of complex tasks.
Mobile devices are personalized computing platforms that allow users to access user-centric services {anywhere} and {anytime}.
Their pervasive nature carries high potential for complex distributed systems across many devices, where multiple users are coordinated to achieve some goal.

These settings can naturally be modelled with Multi-Agent Systems (MAS)~\cite{wooldridge2009}.
MAS support the engineering of complex distributed systems through the decomposition of problems into autonomous agents that perform high-level interactions~\cite{jennings2000,jennings2001}.

Agent-based abstractions can help to build complex systems accross mobile devices~\cite{santi2011}.
Mobile systems can be designed and developed on an elevated abstraction level by making use of high-level agent concepts like \emph{beliefs}, \emph{plans}, and \emph{goals}.
Unfortunately, cognitive agent frameworks~\cite{boissier2013,bordini2007,aschermann2016} lack satisfactory support for the processing of sensor data.
Today's mobile devices are equipped with a rich set of sensors that can be exploited to achieve system behavior that adapts itself to the current situation of the mobile user.
However, the processing of sensor data streams is not an explicitly addressed concern in agent-based systems.

The major problem is the existing \emph{abstraction gap} between data streams and agent percepts~\cite{buford2006,dennis2011,ranathunga2013,ziafati2013}.
Agents use their perception to obtain information from the environment.
The agents expect to receive the information in form of high-level percepts, i.e. the percepts resemble actionable knowledge that the agents can understand and react to directly.
However, sensor data is located on a lower level of abstraction:
The individual data elements of the stream carry little informative value when observed in isolation and cannot be acted upon directly.
To shift the low-level data stream to a higher-level percept stream, meaningful percept patterns have to be detected and encoded into meaningful situations.
For example, a single GPS value in isolation merely allows conclusions about the location of an agent.
Only by analyzing the patterns of multiple data elements, higher-level knowledge can be extracted:
    \emph{Is the agent moving?
    Is it accelerating or slowing down?
    Is it moving faster or slower than other agents?
    }
Without a dedicated stream processing component, agents cannot infer situational knowledge from complex data sequences.
To achieve advanced situation awareness, agents have to analyze the relationships and patterns of multiple data elements and their relations in time.

In a preliminary version of this paper~\cite{doetterl2018at}, we proposed an extension of the BDI architecture called \emph{enhanced perception}. 
This new perception approach includes two new abstractions for the design and development of cognitive agents in mobile ecosystems, namely \emph{expectations} and \emph{interpretations}.
Through enhanced perception, mobile agents~\footnote{We use the term \emph{mobile agent} to refer to an agent that runs on a mobile device.} can perceive higher-level knowledge in low-level streaming data.

In this paper, we present a revised version of our enhanced perception approach, which better reflects the role and influence of beliefs on the agent's perception.
Furthermore, we apply our proposal to a new problem scenario, which highlights the benefits of our approach for agreements between agents:
We present a crowdshipping case study where auctions are used as an agreement mechanism to transfer delivery tasks between agents.
The derived higher-level situations determine \emph{when} auctions are opened, \emph{who} is eligible for participation, and \emph{who} is appointed the winner.
Moreover, we have two implementations to test our proposal:
We have implemented an extension for Jason that puts data stream processing into BDI agents,
as well as a prototype of our case study.
We evaluate our approach via the crowdshipping case study and by running our implementation on real smartphone sensor data.

The rest of the paper is structured as follows.
Section~\ref{sec:related-work} discusses related work.
Section~\ref{sec:concept} introduces our stream-based perception approach.
In section~\ref{sec:case-study}, we perform a case study, which demonstrates the applicability and benefits of our proposal.
The proof-of-concept in section~\ref{sec:proof-of-concept} gives further insights into how our proposal can be applied and reports our experience of integrating the new perception approach into the Jason framework.
Section~\ref{sec:evaluation} evaluates our work with smartphone sensor data.
Finally, section~\ref{sec:conclusion} terminates the paper with conclusions and final remarks.

\section{Related work}
\label{sec:related-work}
There exist different frameworks and middleware for agent-oriented programming on mobile devices~\cite{bergenti2014,santi2011}.
In the past, several agent-based systems have been proposed that operate on mobile devices:
The approaches either present an agent-based mobile system for a specific application domain (e.g., e-health~\cite{chan2008}) or aim to bring agent abstractions to mobile computing platforms \cite{aguero2008,rahwan2004,sartori2013}.
However, none of these approaches uses advanced data processing to detect complex relationships between observed percepts.

Today's agent frameworks focus primarily on beliefs, goals, and plans, which are the core abstractions of agent-based programming.
The processing of percepts, however, is kept simple, which may be sufficient for many use cases but is limiting in sensor environments where data is low-level and has strong temporal relationships.

There are different logic-based proposals that consider temporal aspects.
Allen et al.~\cite{allen1994} describe a temporal logic that gives special consideration to events and actions.
Actions are executed by agents and result in events.
Events are patterns of change that are considered useful and relevant by some agent.
This temporal logic is based on explicit relationships between time periods, such as 
\emph{before},
\emph{meets},
\emph{overlaps},
\emph{during},
etc.
Broda et al.~\cite{broda2009} present SAGE, an environment monitoring system based on logical agents.
SAGE uses forward chaining to detect higher-level events, such as movements, from sensor data.

Dynamic environments are challenging, which has motivated different works on advanced event processing in agents.
Agüero et al.~\cite{aguero2015} present an event processing approach for agents in dynamic environments.
Different event processing strategies are discussed, which result in different agent behaviors.

In current frameworks, plans are usually triggered by the occurrence of a single event.
Buford et al.~\cite{buford2006} extend the BDI agent architecture with event correlation to allow plans to be triggered by a (potentially complex) pattern of multiple events rather than only by a single, isolated event.

Ziafati et al.~\cite{ziafati2013} add advanced event processing to BDI agents in the context of autonomous robot programming.
The robot's sensory information is processed in order to extract relevant knowledge that the robot's control component can use to make and execute appropriate plans.

Dennis et al.~\cite{dennis2011} present a dedicated abstraction engine that enables BDI agents to observe and manage satellite control systems.
This abstraction layer translates from the continuous sensor data produced by the control system to the discrete and abstract information needed by the BDI agent.

There are two approaches related to our work that make use of Complex Event Processing (CEP)~\cite{luckham2001} and the Jason~\cite{bordini2007} agent framework:
Ranathunga et al.~\cite{ranathunga2012} use CEP within a global event processing component to interface Jason with the virtual \emph{Second Life} platform.
Ranathunga and Cranefield~\cite{ranathunga2013} integrate CEP with Jason agents to identify complex situations in the agent's environment.
In both approaches the CEP engine is an external component outside of the BDI agent.

There are some works with particular focus on agent perception.
Cranefield et al.~\cite{cranefield2015} propose the concept of a percept buffer for BDI agents that processes percepts according to customizable policies.
The policies determine how percepts in the buffer should be replaced, merged, or summarized before being made available to the agent's deliberation.
Percept buffers can be used to protect the BDI agents' relatively slow deliberation from the load of high-speed data streams.
Johansson et al.~\cite{johansson2003} propose perception management, which bundles a set of perception-related functionalities and tasks that go beyond mere sensor management.
Weyns et al.~\cite{weyns2004} present a formal model for active perception, which allows agents to direct their attention to the most relevant occurrences in the environment.
Active perception assumes that percepts are given on the knowledge level and narrows the agent's view.
Our enhanced perception approach acknowledges that percepts can appear on a low abstraction level and widens the range of situations the agent can perceive.

Further related work can be found in the area of context-aware systems.
\mbox{Yılmaz} and Erdur \cite{yilmaz2012} present a context-aware MAS where mobile client agents can request context information from a server-side context agent.
To infer the context, the context agent performs rule-based reasoning on a context ontology.
Alfonso-Cendón et al.~\cite{alfonso2016} let agents perform context-aware workflows in the ambient intelligence domain.
The agents run on top of an existing context management system, which serves as the provider of context information. 
We are not aware of any context-aware multi-agent approach where data stream processing is integrated into a cognitive agent architecture to analyze mobile sensor data.

\section{Agents with enhanced perception}
\label{sec:concept}
In this section, we present our approach for cognitive agents operating in mobile ecosystems.

\subsection{Sensing the environment}

Modern smartphones have a wide range of data sources available through which they can capture the immediate environment~\cite{stipkovic2013}:
\begin{itemize}
    \item \emph{Internal sensors}: On-board sensors to measure acceleration, air pressure, GPS, humidity, temperature, rotation, etc.
    \item \emph{External sensors}: Body sensors like bracelets or chest harness that are connected to the smartphone to measure blood pressure or heart rate.
    \item \emph{Other apps}: Any other app on the device. For instance, the calendar app that informs about upcoming appointments.
    \item \emph{Operating system}: Information about system events like low battery state or missed calls.
    \item \emph{Communication interfaces}: Wireless communication, e.g. via WIFI, enables the access of web services or other data sources online.
\end{itemize}

Like Santi et al.~\cite{santi2011}, we consider these data sources artifacts of the environment, where the agent is situated and which the agent can sense to obtain data.
In the A\&A (agents \& artifacts) meta-model~\cite{omicini2008}, artifacts are the passive components of the MAS that are intended to \emph{be used} by the agents.
While agents constitute the pro-active and autonomous components of the system, artifacts can be understood as tools that provide functionality to the agents.
Each artifact exposes a usage interface to the agents, which can consist of arbitrary operations.
Artifacts are a generic mechanism to provide resources to the agents or to provide an interface for performing actions on the environment.

In our approach, we distinguish between two groups of artifacts.
\begin{enumerate}
    \item \emph{Device artifacts} provide access to the different sensors and services of the smartphone.
        Via these artifacts, the agent can obtain sensor data, check sensor availability, activate or deactivate sensors, or change sensor configurations.
    \item \emph{Domain artifacts} provide access to data and functionality that the agent needs to act in the given problem domain.
\end{enumerate}

The artifacts provide the data in form of data streams.
Streaming data has various properties that prevent its direct use in conventional agent architectures.
\begin{itemize}
    \item \emph{Low-quality}~\cite{khaleghi2013}:
        Data streams can be of low quality due to imperfect sensing; data can be missing or inaccurate.
    \item \emph{Low-level}~\cite{ye2012}:
        Due to the existing abstraction gap, it is often not possible to react to single isolated data elements as their meaning is not inherently clear.
\end{itemize}

Other agents can also serve as data sources:
Mobile agents can establish dynamic connections with nearby agents to engage in flexible interaction and data exchange.
The agents communicate via messages.

\subsection{Processing the percept stream}

\subsubsection{Agent architecture}
\begin{figure*}
    \centering
    \includegraphics[width=0.9\textwidth]{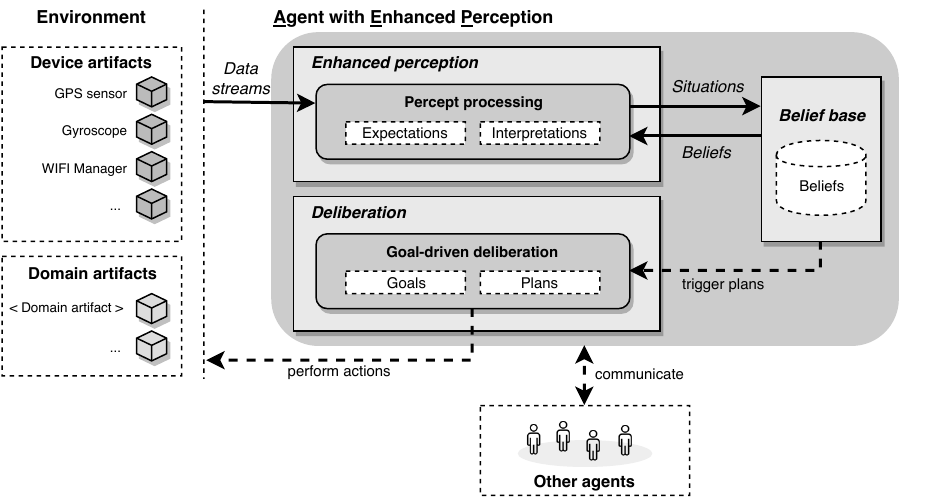}
    \caption{AEP Architecture: \textbf{A}gents with \textbf{E}nhanced \textbf{P}erception}
    \label{fig:enhanced-perception}
\end{figure*}

We propose AEP Architecture (\textbf{A}gents with \textbf{E}nhanced \textbf{P}erception) for agents on mobile devices, as shown in Fig.~\ref{fig:enhanced-perception}.
The AEP architecture consists of three major components:
The \emph{belief base} and \emph{deliberation} component originate in large parts from the conventional agent architecture~\cite{bordini2007,georgeff1987,rao1995}.
\emph{Enhanced perception} constitutes the proposed extension.

\begin{description}
\item[Belief base:]
The belief base holds the agent's beliefs, which are pieces of information about the world that the agent believes to be true.
Beliefs can change through external events in the environment (indicated by percepts), the agent's internal deliberation, or messages received from other agents.
\item[Deliberation:]
The agent deliberates over the world to decide which actions to perform.
The agent's deliberation process is controlled by plans and is aimed at the achievement of the agent's goals.
\end{description}

In mobile environments, these two components are insufficient as the incoming percepts are mostly located on a low abstraction level, which prevents a direct understanding and reaction by the agent.

\begin{description}
    \item[Enhanced perception]
        addresses the challenges introduced by data streams, which allows the agent to detect complex spatio-temporal data patterns and protects the agent's deliberation from low-quality and low-level percepts.
        (First steps towards such an enhanced perception approach were made in \cite{doetterl2017at,doetterl2018at}).
\end{description}

\subsubsection{Exploiting data stream concepts}
As shown in Fig.~\ref{fig:enhanced-perception}, the agent acquires streaming data from device and domain artifacts.
We consider each percept a structure with the following information:
\begin{center}
(type, timestamp, key$_0=$ value$_0$, ..., key$_n=$ value$_n$)
\end{center}

Each percept adheres to a type, which gives the percept a semantic and constrains its admissible key-value pairs.
Examples for percept types are GPS readings, acceleration events, or some significant occurrence in the application domain.
Furthermore, each percept holds the timestamp of its creation, which enables time-based pattern matching over percept sequences.

In the conventional agent architecture, the agent perception processes one percept at a time and neglects the history of recently observed percepts.
To allow agents to perceive complex patterns in percept sequences, we extend the agent's perception by the following information flow concepts~\cite{cugola2012}:
\begin{itemize}
    \item \emph{Selection} of particular percepts that match certain conditions regarding percept type or attribute values.
    \item \emph{Windows} that allow analysis of (i) the last N percept occurrences or (ii) the percept occurrences of the last T time units.
    \item \emph{Aggregates} to combine multiple percepts and the data they carry to new information.
\end{itemize}

Therefore, we process the percept stream with the following information flow operators~\cite{cugola2012,luckham2001}:
\setlength\tabcolsep{0.5em}
\begin{table}[h]
\centering
\begin{tabular}{ r l }
    \emph{p}$_1$ \texttt{and} \emph{p}$_2$ & Conjunction \\
    \emph{p}$_1$ \texttt{->} \emph{p}$_2$ & Sequence (followed by) \\
    \texttt{.window:time(}\emph{t}\texttt{)} & Time window \\
    \texttt{avg(}\emph{x}\texttt{)},
    \texttt{max(}\emph{x}\texttt{)},
    \texttt{min(}\emph{x}\texttt{)}, 
    \texttt{sum(}\emph{x}\texttt{)} 
    & Aggregation \\
\end{tabular}
\end{table}

In AEP, the agent holds expectations and interpretations, which make use of these operators to transform the percept stream.

\subsubsection{Expectations}
\begin{definition}[Expectations]
    An agent's \emph{expectations} characterize the agent's subjective attitude towards percepts. 
    Expectations pose requirements for the data and information carried by the stream of observed percepts.
\end{definition}

Expectations limit the percept stream to those percepts that fulfill the agent's expectations.
Percepts and percept sequences that violate the agent's expectations are purposefully ignored or corrected.
This reduces the agent's computational load, prevents imperfect data manifesting itself in beliefs, and lets the agent gain control over the incoming data.

\begin{example}[Expectation]
    When the user is riding a bike, the agent expects two consequent GPS readings, which are measured within two seconds to each other, to be at most 20 meters apart, which corresponds to a speed of 36km/h.
If the distance is larger than 20 meters, the agent considers one of the data points to be erroneous and drops at least one of them for being inconsistent with the agent's expectations regarding the accuracy of the data.
\end{example}

We follow a rule-based approach, where the detection of a certain pattern in the percept stream triggers the forwarding of the percept instances that fulfill the expectations.
\label{expectationrule}
\begin{Verbatim}[fontsize=\scriptsize,frame=lines,rulecolor=\color{black},framesep=2mm,commandchars=+\{\}]
// "Forward plausible GPS data"
CONDITION:
    (gps1=GPS -> gps2=GPS).window:time(2 seconds)
    where BB.contains("isCycling")
        and Geo.distance(gps1, gps2) < 20 meters
ACTION: forward gps2
\end{Verbatim}
This rule implements the given example.
Event types start with an uppercase letter (e.g., GPS).
Names starting with a lowercase letter are aliases, which are used to refer to a specific event instance (e.g., gps1).
Whenever a GPS event gps1 is followed by a GPS event gps2 within two seconds, the rule checks whether an isCycling belief exists in the agent's belief base (BB).
If this belief exists and the GPS events are within 20 meters of each other, GPS event gps2 is plausible and forwarded to be used in subsequent rules.
BB.contains() is a method invocation that accesses the agent's belief base.
Geo.distance() is a service call to a helper function that returns the distance between two GPS locations.

Expectations enable stream-based and belief-based filtering.
\begin{itemize}
    \item \emph{Stream-based filtering}:
In conventional BDI agents filtering is usually performed by checking each percept in isolation.
Operating on sensor streams, filtering has to go beyond this simple strategy and perform data cleaning by considering multiple data elements and their temporal relationships.
Based on data stream operations, expectations allow implementing advanced cleaning methods that have been proposed in the literature on sensor data streams: to smooth, merge, arbitrate, or virtualize~\cite{jeffery2006,jeffery2006b}.
This can yield more reliable input for the agent's interpretations.

\item \emph{Belief-based filtering}:
An agent's beliefs can provide crucial context for appropriate filtering.
In the example rule above, the agent holds a belief about the user's current mode of transportation (is cycling).
This helps the agent to estimate whether the observed GPS measurements are plausible.
The agent can query the belief base (BB) for the existence of a belief or for concrete values.
        \begin{description}
            \item[BB.contains(\emph{expression})]
                where \emph{expression} may contain placeholders (indicated by an underscore), e.g.:
                BB.contains('temperature(\_)') returns true if the agent holds a belief about some temperature (with an arbitrary temperature value).
            \item[BB.get(\emph{expression})]
            where \emph{expression} contains one unbound variable and arbitrary placeholders, e.g.:
            BB.get('temperature(Var)') returns 20 if the belief base contains the belief temperature(20).
        \end{description}
\end{itemize}

\subsubsection{Interpretations}
\begin{definition}[Interpretations]
An agent's \emph{interpretations} form the agent's rapid recognition capabilities.
Interpretations detect higher-level knowledge in low-level percept streams considering the relationships between multiple percept occurrences.
\end{definition}

Interpretations aggregate and correlate multiple percept occurrences to derive higher-level information.
This higher-level information can either be intermediate \emph{context} or it can be a \emph{situation}.
\begin{itemize}
    \item \emph{Context} is information that is generated by interpretations and that serves as input for other interpretations.
        It is not contained in the incoming data stream directly, but has to be inferred.
        While being more abstract than raw sensor data, in some cases it is still too low-level and too frequent to be processed by the agent's deliberation.
    \item \emph{Situations} are a product of interpretations.
        They are detected by fusion of context information and domain events.
        They are located on a higher abstraction level than context and typically are observed less frequently than context information.
        Situations are meaningful occurrences that are made available to the agent's deliberation in form of beliefs.
\end{itemize}

We give a simple example that derives context from raw sensor data.
\begin{example}[Interpretation]
When the two latest GPS readings are at least one meter apart, the user has moved.
\end{example}

Like expectations, interpretations can be expressed with pattern rules.

\label{interpretationrule}
\begin{Verbatim}[fontsize=\scriptsize,frame=lines,rulecolor=\color{black},framesep=2mm,commandchars=+\{\}]
// "Detect movement"
CONDITION: gps1=GPS -> gps2=GPS
    where Geo.distance(gps1, gps2) > 1 meter
ACTION: create HasMoved
\end{Verbatim}

This rule implements the given example.
When a GPS reading gps1 is followed by a GPS reading gps2 and they are more than one meter apart, a HasMoved context event is created and injected into the data stream.

Interpretation rules enable a stream-based and belief-based detection of situations.
\begin{itemize}
    \item \emph{Stream-based detection}:
        As mentioned earlier, interpretations use data stream concepts (time windows, aggregation, etc.) to detect higher-level information.
    \item \emph{Belief-based detection}:
        Like expectation rules, interpretation rules can access the belief base.
        An agent's belief base can contain important knowledge about the environment, other agents, or the smartphone user.
        For example, an interpretation can only estimate the arrival time at a certain point of interest (another user or some landmark) if its location can be obtained from the belief base.
        Interpretation rules use the same interface as expectation rules, consisting of the methods BB.contains and BB.get.
\end{itemize}

Interpretation rules can perform two different actions.
When a pattern match occurs, they can either (i) create context or a situation and inject it into the data stream for further processing
or (ii) they can add a situation as a belief to the belief base.

\subsubsection{Changing beliefs}
When a new situation is added to the belief base, this new belief might contradict previously held beliefs.
Therefore, some beliefs might have to be removed or updated.
Depending on the concrete characteristics of the application domain, a truth maintenance system might be warranted that keeps the belief base consistent.
In our architecture, we follow the approach used in the Jason framework were the beliefs about the world are maintained by plans or belief annotations that indicate when a belief should expire~\cite[p.~38]{bordini2007}.
Therefore, we consider it primarily a responsibility of the belief base and/or deliberation, which we do not want to focus on in this paper.

\subsection{Taking action}
The addition and deletion of a belief can trigger the execution of plans in the deliberation component of the AEP architecture.
Triggered plans can then initiate actions, e.g., calling an operation provided by an artifact or sending a message to another agent.

\begin{figure*}
    \centering
    \includegraphics[width=0.75\linewidth]{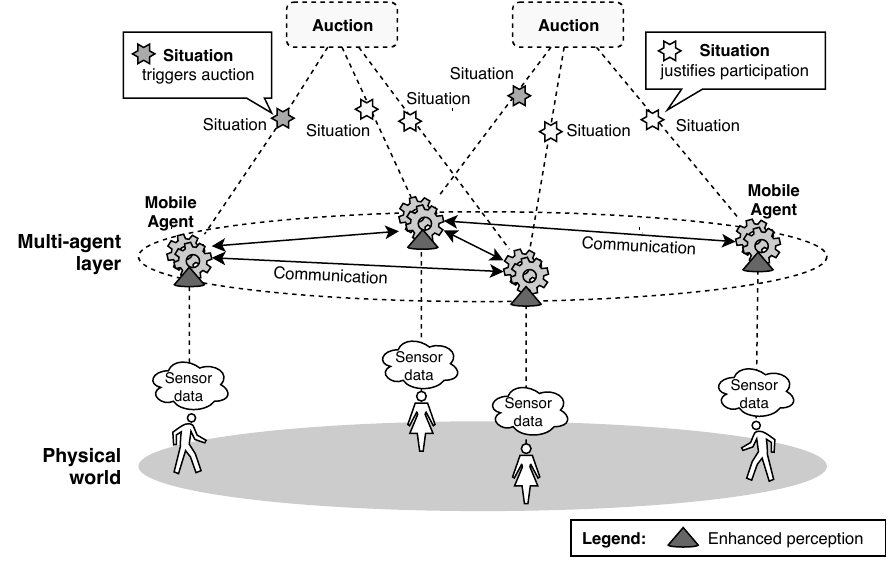}
    \caption{Mobile ecosystem with sensor-driven mobile agents and auction-based agreements}
    \label{fig:mas}
\end{figure*}

In sensor environments, the detected situations are the essential units that trigger agreement processes and underpin the finding of agreements.
Fig.~\ref{fig:mas} illustrates the role of enhanced perception for reaching agreements in distributed mobile systems.
The users' mobile devices sense the physical world and generate sensor data that can be observed by the agents.
Via enhanced perception, the abstraction level of the sensor data is raised by encapsulating meaningful percept sequences into situations.
The situations drive the agents' deliberation and help agents to reach agreements.
Auctions are one possible mechanism how agreements can be reached.
In the upcoming sections, we describe a case study were situations are used to trigger auctions; agents are invited to bid if they satisfy certain situation requirements that justify their participation.

\section{Case study}
\label{sec:case-study}
We present a case study to demonstrate the applicability of our proposal and to illustrate the role of situations in the proposed multi-agent setting.

\subsection{Crowdshipping for the last mile delivery problem}

Nowadays, a large number of customers orders products online, which requires shipping companies to deliver parcels right up to the customers' doorsteps.
Whereas formerly a lot of products could be delivered collectively to shops and sold there, these days many consumers expect their products delivered directly to their individual home addresses.
This large increase in shipping destinations is sometimes referred to as the \emph{last mile delivery problem} \cite{lee2001}.
To overcome this new challenging demand on the shipping companies' delivery infrastructure, industry and academia are working towards new and innovative shipping strategies.

\emph{Crowdshipping} seems a promising solution approach for the last mile delivery problem \cite{marcucci2017}.
Crowdshipping is based on the observation that private citizens in their daily lives walk, ride and drive on the same roads and routes that the parcels have to take to be delivered to the recipients' homes.
The delivery capacity could be substantially increased if private citizens carried parcels towards their destinations with little or no detours in exchange for monetary compensation.

Recent publications on crowdshipping investigate assignment procedures that match the available crowdworkers with delivery tasks or perform route planning to minimize the length of the workers' shipping routes \cite{wang2016,archetti2016,dayarian2017,chen2018,arslan2019}.
Until now, only few proposals have been made that use the multi-agent paradigm and that explicitly enable or encourage the transfer of parcels between workers \cite{giret2018,rodriguez2018}.
In the subsequent sections, we present a multi-agent crowdshipping system where the workers have the autonomy to hand over parcels to other workers who then continue the delivery.
This way, delivery chains can form that carry parcels to their destinations.
In contrast to other works, we employ data stream processing on the workers' smartphones to incorporate knowledge about the current situation into the crowdshipping system.

\subsection{Dynamic crowdshipping with AEP}

We describe a multi-agent crowdshipping system that benefits from the enhanced perception approach of the previous section.

\paragraph{General approach}
The crowdshipping MAS consists of the following entities:
\begin{description}
    \item[Parcels] are packages that have to be delivered to (or towards) a certain destination.
        Each parcel has a pre-determined payout value that is rewarded to the user who delivers the parcel to its final destination.
    \item[Users] are private citizens that have agreed to participate as shippers in exchange for financial rewards.
        Users can decide whether to deliver a parcel to its destination and receive the full reward or to cover only a part of the delivery distance and sell the parcel to another user who performs the rest of the delivery (or sells it to another user).
    \item[User agents (UAs)] support users in their decision making.
        The UAs run on the users' personal smartphones and monitor the environment to perform situation-aware actions, including \emph{obtaining tasks} and \emph{selling tasks} while considering the user's situation and financial interests.
\end{description}

To sell and buy tasks, the MAS uses an auction mechanism with situation-dependent parameters.
Auctions are an important agreement concept to allocate resources (goods, tasks, etc.) to self-interested agents in a multi-agent setting \cite{shoham2009}.
One agent acts as the \emph{auctioneer} and collects the bids from interested buyers.
The auctioneer then assigns the resources according to some clearance rule and collects the payments of the bidders according to some payment rule.
In this case study, we do not focus on optimal winner determination or payment rules.
We rather try to highlight how situations derived from data streams can control and influence agent interaction to come to agreements.

\begin{itemize}
    \item
        \emph{Opening auctions}:
        The UA opens an auction whenever a user is holding a parcel and the user is in a situation where transferring the parcel to another user is preferable.
        \textbf{Situation}: {The user has not been making any delivery progress in the last 5 minutes.}
    \item
        \emph{Selecting participants}:
        When the UA opens an auction to receive bids from other UAs, it announces the auction to a selected set of participants.
        A call for bids is explicitly addressed to agents that have detected a particular user situation.
        \textbf{Situation}: The user can arrive to pick up the parcel within five minutes.
        (Whether a user can arrive within a certain time window depends on the user's physical distance, mode of transportation, and movement behavior of the recent past.)
    \item
        \emph{Determining winners}:
        When the UA has received all bids, it applies a situation-dependent strategy to determine the winner of the auction.
        For example, the UA might not necessarily declare the highest bid the winner if another agent exists that has bid a similar amount but can arrive to pick up the parcel three minutes faster.
        \footnote{Agents would have to be penalized if they report situations inaccurately or maliciously.}
\end{itemize}

\begin{figure*}
    \centering
    \includegraphics[width=0.85\linewidth]{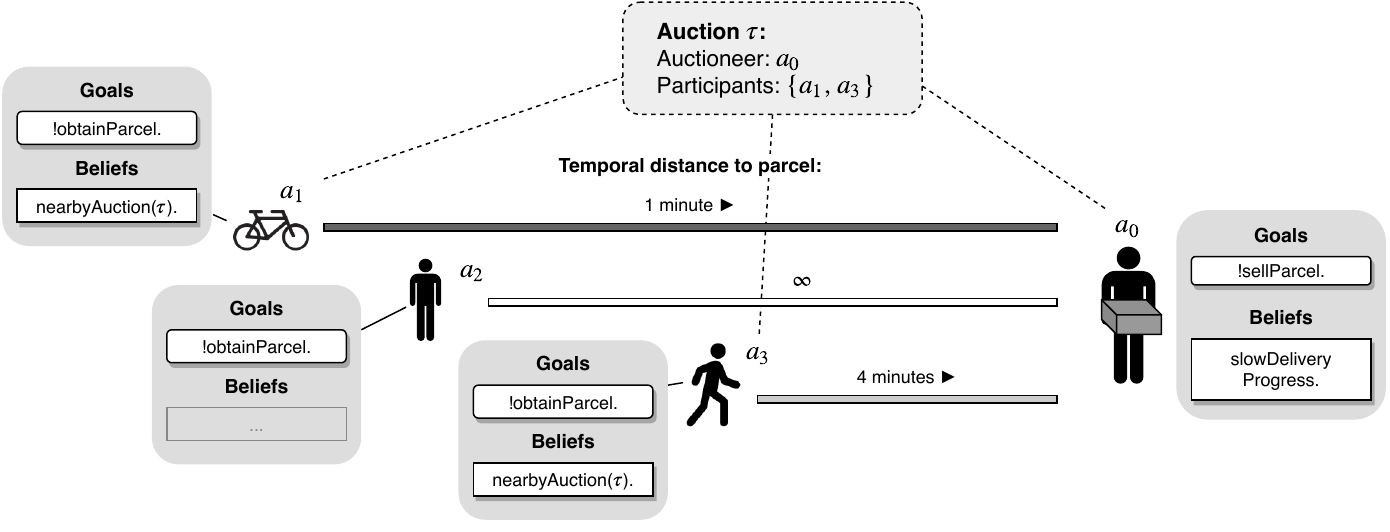}
    \caption{Multi-agent crowdshipping scenario}
    \label{fig:case-study-process}
\end{figure*}

\paragraph{Example}
Fig.~\ref{fig:case-study-process} shows an exemplary scenario, which consists of four agents $a_0$ to $a_4$, whereby one of the agents ($a_0$) is currently holding a parcel.
The agents compute their local situations and adapt their beliefs and goals accordingly.
Thereby, they rely on their enhanced perception component.
Using its interpretation capabilities, agent $a_0$ has detected that its user has not made any delivery progress recently and therefore adopts the goal \emph{!sellParcel}.
This triggers the auction mechanism.
The agents $a_1$, $a_2$, and $a_3$ get notified about the nearby auction and check whether they fulfill the participation requirement.
Agent $a_2$ has not been moving in the recent past (the user might be busy or resting) and hence does not participate in the auction.
Agents $a_1$ and $a_3$ both estimate a possible arrival time at agent $a_0$ in below 5 minutes and therefore consider participation.
Using their beliefs and plans, they check whether the parcel's destination is in accordance with the user's movement direction.
As this is the case for $a_1$ and $a_3$, they both decide to submit a bid to $a_0$.
Agent $a_0$ collects the bids and determines the winner.
Having bid more, the parcel is assigned to agent $a_1$ who not only is willing to perform the task for less revenue (the revenue is the parcel reward minus the bid made in this auction), but also can arrive faster to pick up the parcel.

\paragraph{Application of concepts}
The conceptual building blocks of AEP can be applied as follows to implement the crowdshipping case study.
The agent obtains GPS data from the GPS sensor artifact, which is processed by the enhanced perception component.
\begin{description}
    \item[Expectations] clean the GPS sensor data by removing implausible GPS readings.
        Therefore, they consult the belief base to better anticipate the characteristics of correct GPS readings:
        when the user is riding a bike or a car, GPS events may lie farther apart than when the user is walking.
    \item[Interpretations] detect the higher-level situations that drive the auction mechanism. Among others, they
                detect slow delivery progress
                and
                estimate arrival times at nearby auctions/parcels.
        These are meaningful occurrences that are not inherent in a single individual GPS event but can be derived from sequences of GPS events.
    \item[Beliefs] represent an agent's current knowledge about the crowdshipping system, including
                    the user's state (has parcel / has no parcel),
                    the user's financial profile and walking preferences (do not accept tasks with rewards below 2.00 EURO, do not accept tasks farther away than 500 meters), etc.
        Beliefs are used in expectations, interpretations, and plans; they influence the agent's percept processing and deliberation.
    \item[Goals] express the agent's current attitude towards parcels and the agent's ambitions in the auction process.
        The goals include: sell parcel, obtain parcel, and participate in auction.
    \item[Plans] get triggered when beliefs change or goals get adopted.
        Most importantly, the agent holds plans that allow it to open auctions and to participate in auctions that other agents announce.
        These plans implement the agent's winner determination strategy as an auctioneer
        and the agent's bidding strategy as a participant in another agent's auction.
\end{description}

The case study shows that the crowdshipping scenario can be naturally expressed using MAS:
each user of the crowd is represented by a self-interested agent; the decision making is distributed over the users' mobile devices.
Using enhanced perception, the agent interaction (in form of auctions) is driven by higher-level situations that are derived from lower-level sensor data.
Interpretations allow agents to perceive complex spatio-temporal patterns across a stream of low-level percepts, e.g., slow delivery progress in a GPS stream.

\section{Proof of concept}
\label{sec:proof-of-concept}
In this section, we show some of the most important interpretation rules that are needed to implement the multi-agent crowdshipping system as outlined in the previous section.
We also describe our technical integration of enhanced perception into the Jason framework and discuss the most important benefits of our extension compared to Jason agents without enhanced perception.

\subsection{Interpretation rules for the multi-agent crowdshipping system}
\label{sec:proof-of-concept-rules}
\begin{figure}
    \centering
    \includegraphics[width=1.00\linewidth]{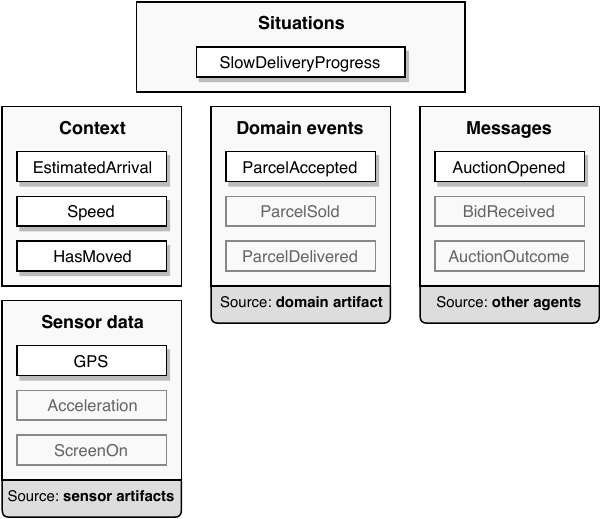}
    \caption{Events in the multi-agent crowdshipping system}
    \label{fig:case-study-events}
\end{figure}

We have a look at the following interpretation rules (R.1 and R.2).
\begin{description}
    \item[R.1] (Agent with parcel): When user holds parcel and makes slow delivery progress, try to sell the parcel by opening an auction to nearby agents.
    \item[R.2] (Agent without parcel): When an auction was opened nearby, check whether the auctioneer can be arrived within the required time limit. If this is the case, consider participating in the auction by sending a bid.
\end{description}
In the interpretation rules we make use of the event types shown in Fig.~\ref{fig:case-study-events}.
There are three categories of event types that can be observed by the agent.
\begin{description}
    \item[Sensor data] is accessed from the smartphone sensors via the sensor artifacts.
        Event types of this category are GPS readings, acceleration data, or whether the smartphone's screen is active.
        Here we focus on GPS data.
    \item[Domain events] stem from the domain artifact and indicate significant occurrences in the crowdshipping domain.
        Here, an agent receives explicit notification when a parcel was accepted (either received from the shipping company or bought from another agent).
    \item[Messages] are announcements made by other agents.
        In our setting, agents use messages to inform nearby agents about new auctions, bids, and auction outcomes.
\end{description}
While the event types of the category of domain events have to be tailored to the specific problem domain, the event types of the category sensor data are more general and reusable for a wide range of problem domains. 

\textbf{Context} that can be derived is the user's movement, speed, and estimated arrival time at nearby auctions.
Based on context, domain events, and the content of messages, \textbf{situations} such as slow delivery progress can be detected.

To express the interpretation rules, we use a declarative Event Processing Language (EPL).
There are different families of EPL.
Here, we use a syntax that is inspired by SQL and close to the syntax used by the Esper engine.

\subsubsection{Opening the auction (Rule R.1)}

The agent opens an auction when it is making little or no progress in delivering the parcel to its destination.
To detect this higher-level situation, two interpretation rules are used.
The first rule (R.1a) continuously computes the user's distance to the parcel destination.
The second rule (R.1b) then uses this distance to detect when the distance does not become smaller over time.

\begin{Verbatim}[fontsize=\scriptsize,frame=lines,rulecolor=\color{black},framesep=2mm,commandchars=\#\{\}]
// R.1a: "Compute distance to destination"
SELECT Geo.distance(
    move.lat,
    move.lon,
    BB.get("hasParcel(" + parcel.id + ", Lat, _)"),
    BB.get("hasParcel(" + parcel.id + ", _, Lon)")
    ) as #underline{distanceToDestination}
FROM pattern[
  (every parcel=ParcelAccepted)
          -> (every move=HasMoved)
]
WHERE
  BB.contains("hasParcel(" + parcel.id + ", _, _)")
ACTION:
  new DistanceToDestination(#underline{distanceToDestination})
\end{Verbatim}
The rule consists of two parts:
The first part describes a percept pattern using the SQL-inspired SELECT-FROM-WHERE structure.
The second part specifies the action that is executed when a pattern match occurs.
Rule R.1a matches every time a ParcelAccepted event is followed by a HasMoved event and the belief base contains the belief that the agent is still in possession of this parcel.
(If the user had sold or delivered the parcel in the meantime, the belief would have been removed accordingly.)
Whenever this is the case, the distance of the user's current position (move.lat, move.lon) to the parcel's destination (retrieved from the belief base) is computed and encoded into a new event.
The resulting DistanceToDestination event is pushed into the percept stream.

\begin{Verbatim}[fontsize=\scriptsize,frame=lines,rulecolor=\color{black},framesep=2mm,commandchars=\#\{\},numbers=left]
// R.1b: "Detect slow delivery progress"
SELECT #underline{parcel}
FROM pattern [
  every parcel=ParcelAccepted
  ->
  (
    (timer:interval(5 minutes) and not HasMoved)
  or
    (d1=DistanceToDestination(parcelId=parcel.id)
    ->
    d2=DistanceToDestination(parcelId=parcel.id,
             timestamp > d1.timestamp + 5 minutes)
    )
  )
]
WHERE 
  BB.contains("hasParcel(" + parcel.id + ", _, _)")
  and d2.value > d1.value - 20 meters
ACTION:
  BB.add("SlowDeliveryProgress(" + #underline{parcel} + ")")
\end{Verbatim}
\label{rule1b}
When the user accepts a parcel (line 4) and as long as the user is still carrying this parcel (indicated by a corresponding belief; line 17), the following two cases are checked:
\begin{itemize}
    \item
Line 7: Either the user has not been moving for 5 minutes.
    \item
Lines 9-12 and 18: Or, comparing to distance events that appear at least 5 minutes apart, the user has reduced the distance to the destination by less than 20 meters.
\end{itemize}
In either of the two cases, the UA detects that the delivery progress is insufficient and inserts a corresponding belief into the belief base.

The UA has a plan that describes the action that is to be taken in this situation.
In this plan, the UA broadcasts to nearby agents that it has opened an auction to sell the given parcel.
This invitation to the auction includes a description of the parcel, a description which agents are eligible to participate in the auction (e.g., all users who can pick up the parcel within 5 minutes), the current location of the parcel, and a bidding interval that denotes how much time the agents have to announce their bids.

\subsubsection{Deciding whether to participate (Rule R.2)}
Next, the nearby agents receive the invitation to the auction and have to decide whether they participate.
Therefore, the agents check whether they fulfill the participation requirement, i.e. whether the user's recent movement behavior indicates that the parcel could be reached within 5 minutes.

We accomplish this in two steps:
Estimating the arrival time at the parcel (R.2a) and comparing it with the announced pickup deadline (R.2b).
\begin{Verbatim}[fontsize=\scriptsize,frame=lines,rulecolor=\color{black},framesep=2mm,commandchars=\#\{\}]
// R.2a: "Estimate arrival time at auction"
SELECT  auction.id as #underline{AID},
        Geo.distance(auction.location,
                loc.location) as #underline{distance},
        speed.value as #underline{speed}
FROM pattern [
        every auction=AuctionOpened
        and loc=UserLocation
        and speed=Speed
        and speed.value > 0
]
ACTION: new EstimatedArrival(#underline{AID}, #underline{distance}/#underline{speed})
\end{Verbatim}
The arrival time is estimated as follows:
Whenever the agent is notified that an auction was opened, it gathers its most recent data about the user's location and speed.
If the speed is larger than 0, the distance to the parcel location is computed and divided by the speed value.

\begin{Verbatim}[fontsize=\scriptsize,frame=lines,rulecolor=\color{black},framesep=2mm,commandchars=\#\{\}]
// R.2b: "Consider bidding if auction can be
//      reached in time"
SELECT #underline{auction}
FROM pattern [
  every auction=AuctionOpened
  ->
  arrival=EstimatedArrival
      .window:time(auction.biddingInterval)
    and arrival.auctionId=auction.id
    and arrival.estimatedArrivalTime
        < auction.pickupDeadline
]
ACTION:
  BB.add("nearbyAuction(" + #underline{auction} + ")")
\end{Verbatim}
When an auction was opened and an arrival time was estimated within the announced bidding interval, the rule checks whether the estimated arrival time is smaller than the pickup deadline.
If this is the case, the agent adds the new belief \emph{nearbyAuction} to its belief base.

In a corresponding plan that reacts to this belief addition, the agent checks whether the location and destination of the parcel lie on the user's route.
Assuming that this is the case, the agent has decide on the exact value it wants to bid.
This decision depends on the user's monetary profile and the detour that has to be made for participation in the shipping of the parcel.
Having decided on a value, the agent sends its bid to the auctioneer.
The auctioneer gathers all bids and determines the winner.
The auctioneer then notifies the participants about the auction outcome and waits for the winner to come by and pick up the parcel.

\subsection{Integrating enhanced perception into Jason}

To test our approach, we integrated the enhanced perception component into the Jason agent framework.
We rely on Jason's deliberation and belief base and extend it by our customized perception.
Our technical solution is shown in Fig.~\ref{fig:implementation}:
\begin{figure}
   \includegraphics[width=1.0\linewidth]{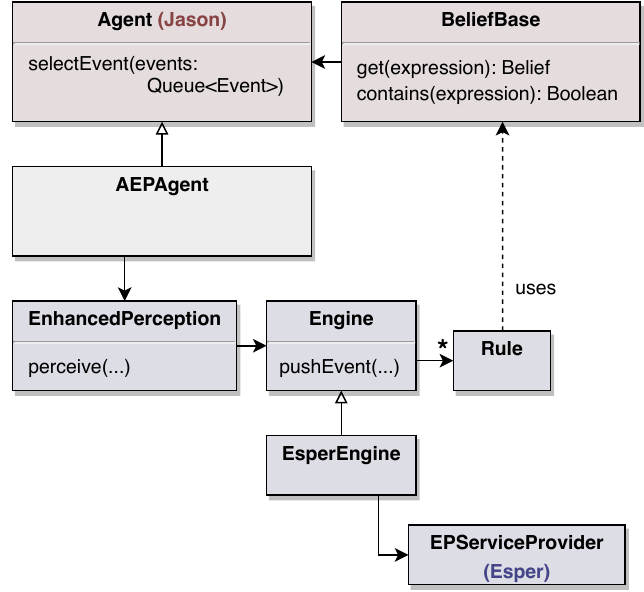}
    \caption{Technical integration
    }
    \label{fig:implementation}
\end{figure}
We developed a new agent class \texttt{AEPAgent}, which is a subclass of the Jason \texttt{Agent} class. 
We override the selectEvent method such that in each processing cycle all percepts are removed from the agent's event queue and passed to the enhanced perception component.
The \texttt{EnhancedPerception} class converts the Jason literals into the event format expected by the event processing engine and pushes the event into the engine.
The class \texttt{EsperEngine} is a wrapper around the Esper CEP engine~\footnote{https://www.espertech.com/esper/ (Accessed: 2019-01-15)}; the class \texttt{EPServiceProvider} is the entry point into the Esper API.
Each engine holds an arbitrary number of rules, which perform the percept processing and have access to the \texttt{BeliefBase} class.
The belief base class provides read and write access to the Jason belief base (via the Jason Agent class).
Whenever a rule infers a situation, the situation is encoded into a Jason literal and inserted into the Jason belief base.
The belief is then available within the agent code, which is written in AgentSpeak and stored in *.asl files (not shown in the figure).
The belief insertion can trigger plans, which can perform actions or send messages to other agents.

\subsection{Implementing the case study with AEP}

We implemented the multi-agent crowdshipping case study with our framework.
We developed a user agent that consists of the perception rules presented in section~\ref{sec:proof-of-concept-rules}.
The user agent obtains a GPS data stream from a GPS artifact.
The GPS data arrives in form of percepts that the agent processes with its enhanced perception module.

The agent's plans are developed in AgentSpeak.
The plans are triggered by belief changes and situations that are produced by the enhanced perception module.

Using our approach, we observe the following benefits.
\subsubsection{Complex spatio-temporal pattern matching}
    By integrating data stream processing into BDI agents, an agent can perform pattern matching on percept sequences.
    In conventional BDI agents, plans are triggered by single events.
    AEP on the other hand enables the triggering of plans by multiple events that can be in complex temporal relationships.
    This capability is necessary when the individual data elements are not meaningful in isolation, but expose their meaning if perceived together.
    In our case study, a single GPS event does not carry enough meaning to trigger the auction plans.
    Only the situations derived from the GPS stream give enough insights whether to invoke the plans.

    Advanced pattern matching over percept streams cannot be easily implemented with AgentSpeak.
    AgentSpeak is well suited to express plans for intelligent behavior on the domain level.
    However, not all problems present themselves at this abstraction level directly: on smartphones, sensors provide data on a lower level of abstraction where the individual data elements in isolation have little meaning.
    If we had to implement the percept processing with plans, we had to attempt this without many essential data stream concepts, most importantly
        temporal windows, non-appearance of events, and aggregations over windows.

        As an example, rule R.1b (section~\ref{rule1b}) triggers when after the acceptance of a parcel 5 minutes pass without the occurrence of a HasMoved event.
This is impossible to implement solely with AgentSpeak plans, as in Jason plans are triggered when certain events \emph{do appear}.
However, there is no mechanism that causes plans to be triggered by events that \emph{do not appear}.
In our approach such situations can indeed be detected by the agent, as illustrated by rule R.1b.

\subsubsection{Reduced deliberation efforts}
    In sensor environments, a conventional BDI agent is confronted with many percepts that it has to reason about.
    The environment appears dynamic and many plans get triggered that alter beliefs and goals.
    In AEP architecture, the agent's expectations and interpretations face this dynamic and pass only filtered, aggregated, and meaningful situations to the deliberation component.
    To the deliberation component, the environment manifests itself less dynamic.
    This fits well the metaphor of the BDI model which is inspired by the human mind:
        the human perception pays attention to only some of the expressions it gathers and
        detects meaningful patterns.
    This is not a conscious effort that is performed by deliberating over the individual stimuli.

\section{Experimental evaluation}
\label{sec:evaluation}

The runtime behavior of our proposal can best be demonstrated by confronting it with real smartphone sensor data.

\begin{figure*}
\centering
\begin{subfigure}{0.32\textwidth}
    \centering
    \includegraphics[width=0.8\textwidth]{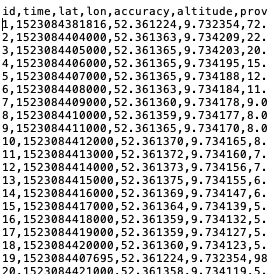}
    \caption{Unprocessed GPS events}
    \label{fig:gps-raw}
\end{subfigure}%
\begin{subfigure}{0.32\textwidth}
    \centering
    \includegraphics[width=0.8\textwidth]{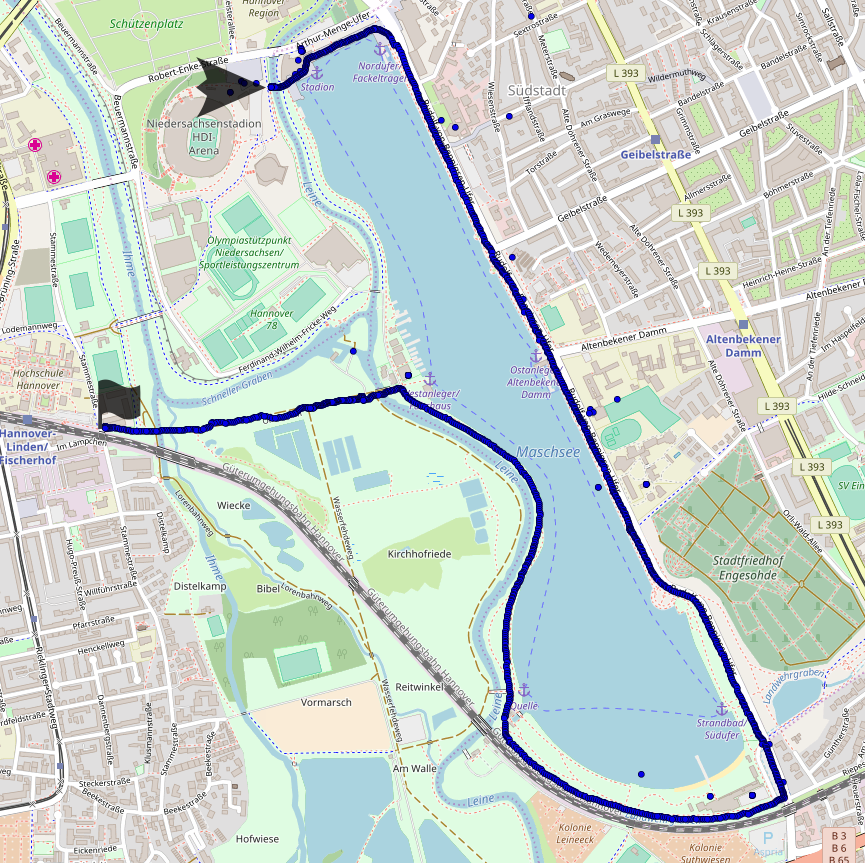}
    \caption{Route taken by the cyclist}
    \label{fig:gps-route}
\end{subfigure}%
\begin{subfigure}{0.32\textwidth}
    \centering
    \includegraphics[width=0.9\textwidth]{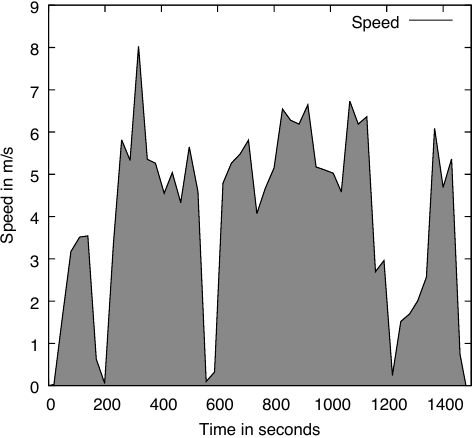}
    \caption{Speed of the cyclist}
    \label{fig:gps-speed}
\end{subfigure}
\caption{GPS data set}
\end{figure*}

\paragraph{GPS data set}
We obtained a GPS data stream on an Android device using Android's LocationManager API for fine-grained location tracking (see Fig.~\ref{fig:gps-raw}).
During 25 minutes, the user performed a short trip using a bike (see Fig.~\ref{fig:gps-route}).
The data set contains three breaks where the user stopped the bike and one phase of increased speed where the user reached more than 7 m/s.
Fig.~\ref{fig:gps-speed} shows the user's speed during the 25-minute period.
In total, 1568 GPS events were captured.

\paragraph{Procedure}
We expose our agent to a replay of the captured GPS stream.
Additionally, we inject three synthetic domains into the stream: two events indicating that a parcel was accepted (ParcelAccepted) and one that indicates that the parcel was delivered (ParcelDelivered).
We log the intermediate context and high-level situations generated by the enhanced perception component.

\begin{figure*}
    \centering
    \includegraphics[width=1.0\textwidth]{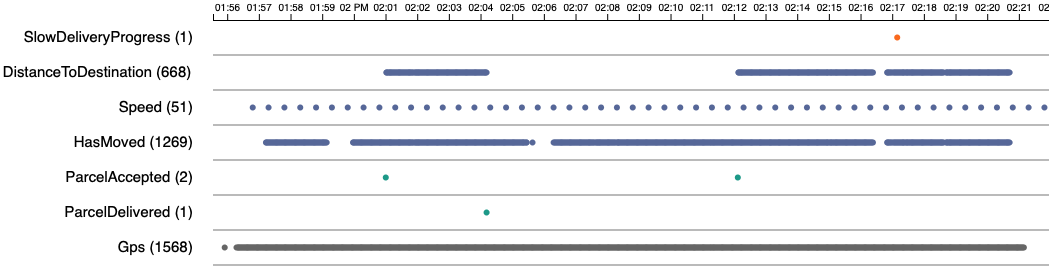}
    \caption{Event series of sensor events (GPS),
    synthetic domain events (ParcelAccepted, ParcelDelivered),
    context (HasMoved, Speed, DistanceToDestination), and
    derived higher-level situations (SlowDeliveryProgress)
    }
    \label{fig:case-study-event-series}
\end{figure*}

\paragraph{Results}
The results are listed in Table~\ref{tab:events} and visualized in Fig.~\ref{fig:case-study-event-series}.
\begin{table}
\centering
\footnotesize
    \caption{Events that were sensed, generated, and inferred during a time span of 25 minutes}
\begin{tabular}{@{}lllr@{}} \toprule
    \textbf{Event} & \textbf{Category} & \textbf{Source} & \textbf{Count} \\ \midrule
    Gps & Sensor data & GPS sensor & 1568 \\
    \midrule
    ParcelAccepted & Domain event &  Synthetic & 2 \\
    ParcelDelivered & Domain event & Synthetic & 1 \\
    \midrule
    HasMoved & Context & Interpretation & 1269 \\
    Speed & Context & Interpretation & 51 \\
    DistanceToDestination & Context & Interpretation & 668 \\
    \midrule
    SlowDeliveryProgress & Situation & Interpretation & 1 \\
\bottomrule
\end{tabular}
\label{tab:events}
\end{table}

\paragraph{Discussion}
Table~\ref{tab:events} lists the observed events, grouped by their respective category.
In total, the agent acquires 1571 data elements from the environment (1568 GPS events and 3 domain events).
From this raw data, the agent infers (1269+51+668=) 1988 intermediate context events.
In the analyzed setting, a single slow delivery situation was detected.
The numbers support our argument that the environment appears less dynamic for the agent's deliberation: the deliberation only receives the slow delivery notification.
Depending on the interpretations, a large number of intermediate context events might be generated that even exceeds the number of sensor events.
However, this is mostly not a problem as long as the load keeps reasonably low - modern event processing engines are quite efficient - and as long as these context events are not pushed into the belief base.

A visualization of the events over time (see Fig.~\ref{fig:case-study-event-series}) confirms the correct behavior of our interpretation rules.
Each event is represented by a circle; when they occur in small intervals they appear to be a straight line.
Having a look at the HasMoved events, we can clearly observe the three breaks that are also apparent in Fig.~\ref{fig:gps-speed}.
The Speed events appear in a 30 second interval; this is due to the use of a batch window of 30 seconds\footnote{The corresponding interpretation rule is omitted in this paper.}.
If we have a look at the DistanceToDestination events, two observations can be made:
Firstly, these events are only generated between the acceptance of a parcel (ParcelAccepted event) and the delivery (ParcelDelivered event) or selling of a parcel.
Secondly, the distance is only re-computed when the user has moved.
Both observations correspond with the desired behavior.
Finally, one slow delivery event is detected.

\section{Conclusion}
\label{sec:conclusion}

In this paper, we have put forward an enhanced perception approach to put advanced data stream processing into cognitive agents in mobile ecosystems.
Enhanced perception is driven by the agent's expectations and interpretations, which integrate information flow concepts into the agent's perceive-deliberate-think cycle.
Expectations and interpretations have access to the agent's belief base, which enables the detection of situations by fusion of
(i) low-level, fast-paced sensor data and (ii) high-level, semi-static beliefs.

We have evaluated our approach by different means.
We applied AEP architecture to a scenario in the crowdshipping domain, where multiple self-interested agents obtain, sell, and deliver parcels to gain financial rewards.
This case study demonstrates the applicability of our proposal and illustrates the role of derived situations for agreements between agents:
the crowdshipping system uses auctions that are initiated and influenced by the situations produced by the agents' interpretations.

We have shown and discussed concrete interpretation rules and the benefits they bring to the conventional BDI-based agent systems:
Most importantly, they enable complex spatio-temporal pattern matching over percepts and relieve the agents' deliberation in dynamic sensor environments.

Our conceptual work is backed by two implementations.
We integrated the Esper CEP engine into the Jason agent framework, which enriches Jason agents with data stream processing capabilities.
We then used this extension to implement parts of the crowdshipping case study.

Finally, we confronted our implementation with real smartphone sensor data.
Using enhanced perception, the agent accurately encodes the GPS data into meaningful context and situations and effectively reduces the number of events affecting the deliberation component.
The deliberation component operates on a reasonable number of meaningful beliefs rather than a high load of low-level, individually meaningless data stream elements.

Future work could investigate how multiple agents can cooperate to detect complex composite situations that a single agent cannot detect on its own due to missing data or interpretations.
In this case it could be helpful to perform parts of the data processing tasks outside of the agents perception in a dedicated data stream processing artifact that all collaborating agents can access.
Furthermore, perception rules do not necessarily have to be explicitly programmed by an agent designer with domain knowledge.
Future work could include a learning approach that derives perception rules from previously observed percept streams and situations.

\nocite{label} 
\bibliographystyle{ios1}           
\bibliography{bibliography}        

\end{document}